\documentclass[10pt,twocolumn,letterpaper]{article}

\usepackage{cvpr}              
\usepackage{booktabs}

\definecolor{cvprblue}{rgb}{0.21,0.49,0.74}
\usepackage[pagebackref,breaklinks,colorlinks,allcolors=cvprblue]{hyperref}

\def\paperID{4534} 
\def\confName{CVPR}
\def\confYear{2025}

\title{LATTE-MV: Learning to Anticipate Table Tennis Hits from Monocular Videos}

\author{
    Daniel Etaat \quad
    Dvij Kalaria \quad
    Nima Rahmanian \quad
    S. Shankar Sastry \\
    University of California, Berkeley \\
    {\tt\small \{danielbetaat, dvijk, rahmanian, sastry\}@berkeley.edu}
}

\begin{document}
\maketitle
\begin{abstract}

Physical agility is a necessary skill in competitive table tennis, but by no means sufficient. Champions excel in this fast-paced and highly dynamic environment by anticipating their opponent’s intent – buying themselves the necessary time to react. In this work, we take one step towards designing such an anticipatory agent. Previous works have developed systems capable of real-time table tennis gameplay, though they often do not leverage anticipation. Among the works that forecast opponent actions, their approaches are limited by dataset size and variety. Our paper contributes (1) a scalable system for reconstructing monocular video of table tennis matches in 3D and (2) an uncertainty-aware controller that anticipates opponent actions. We demonstrate in simulation that our policy improves the ball return rate against high-speed hits from 49.9\% to 59.0\% as compared to a baseline non-anticipatory policy. 

\hspace{-5mm} \textbf{Project website}: \href{https://sastry-group.github.io/LATTE-MV/}{https://sastry-group.github.io/LATTE-MV/}

\end{abstract}

\vspace{-3mm}
\section{Introduction}
\vspace{-1mm}

Table tennis has served as a testbed for robotics research since the 1980s~\cite{robotpingpong980}, presenting unique challenges due to the game's high speed and dynamic environment. The sport requires targeted perception, precise motor control, and strategic decision-making, pushing the boundaries of robotic capabilities in sensing, actuation, and real-time processing.

Recent advancements in deep learning have propelled the development of robotic systems capable of playing table tennis~\cite{isim2real,google,sampleEfficientRL,Gao2020,Huang2015,Ding2022}. These systems have achieved milestones such as cooperative rallying with humans~\cite{isim2real} and competitive play~\cite{google}. Despite these achievements, current state-of-the-art systems struggle against professional players in fast-paced rallies. One explanation for this shortcoming is that the state-of-the-art systems are unable to anticipate their opponent’s future intent when returning high-speed hits.

\begin{figure}[h!]
    \centering
    \includegraphics[width=\linewidth]{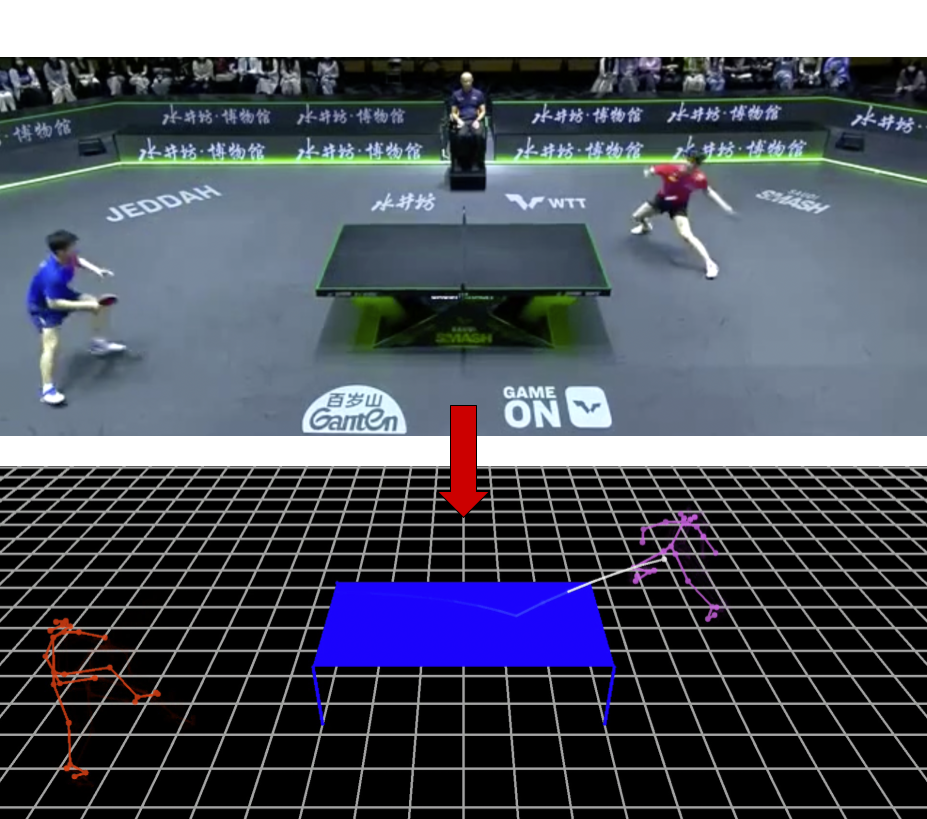}
    \caption{Example of a reconstructed frame.}
    \label{fig:reconstruction_example}
    \vspace{-5mm}    
\end{figure}

Human-action anticipation has been widely studied in robotics ~\cite{Hauser2012,Kuderer2012,Wang2013} and sports~\cite{Daiki2018,Sato2019,Wu2019}, where it is often regarded as a distinguishing characteristic between amateur and expert athletes ~\cite{Daiki2018}. Despite these findings, anticipation is yet to play a prominent role in strategic algorithms for table tennis robotics. Previous efforts to forecast opponent serves~\cite{futurepong} and incorporate anticipatory action selection~\cite{janpeters, prob_modeling_of_human_movements} have been constrained by small datasets. Access to a large-scale, high-quality dataset of competitive gameplay, a resource currently unavailable, may streamline the development of behavior-aware learning-based algorithms -- such as an anticipatory mechanism. 











To address this gap, we introduce LATTE-MV (\textbf{L}earning \textbf{A}nticipation for \textbf{T}able \textbf{T}ennis \textbf{E}xecution from \textbf{M}onocular \textbf{V}ideos) that primarily contributes a novel system for reconstructing monocular video of table tennis gameplay in 3D. By using monocular videos as input, our system scales to the wealth of publicly available footage of competitive gameplay. To the best of our knowledge, this work marks the first attempt at collecting a dataset of table tennis gameplay at scale without specialized recording equipment.
We process $\sim$50 hours of table tennis gameplay into 73,222 exchanges, surpassing the size of the largest collected dataset publicly referenced (14,241 segments in~\cite{google}). We note that this is not a fair comparison, though, since our reconstruction algorithm does not recover racket poses and our reconstructions are likely less accurate. We also note that our data is composed of human-human gameplay featuring \textit{professional} athletes, as opposed to human-robot gameplay or amateur human-human gameplay.


With this dataset in hand, we train a transformer-based generative model \cite{transformer} to learn the dynamics of table tennis gameplay. This model serves as a foundation for anticipatory action prediction, allowing us to forecast an opponent's future shots based on their prior actions. We further reason about epistemic uncertainty in the model's prediction through conformal bounds which can be used to filter out uncertain predictions. 
We demonstrate in simulation that our model's predictions can be used to improve the ball return rate of a KUKA robot arm from 49.9\% to 59.0\% as compared to a baseline non-anticipatory policy.

In conclusion, we make the following contributions: 1) we unify the use of various pre-trained models and develop a novel reconstruction system to extract expert gameplay data from large-scale open-world videos; 2) we apply this dataset to learn an uncertainty-aware anticipatory control algorithm. We deploy our controller on a simulated robot table tennis player and show that it improves the ball return rate from \textbf{49.9\%} to \textbf{59.0\%}.
\section{Related Works}

\noindent \textbf{Reconstruction of Sports Gameplay from Video}\,\,\,\, 
Reconstructing sports footage is a common area of research within the computer vision community. Some works focus on entity tracking and segmentation on 2D images. ~\cite{ttnet} trains a neural network to perform real-time player, ball, and table segmentation and~\cite{tracknetv3} trains a network for high-speed ball tracking. General purpose pre-trained object detection and image segmentation models like YOLO~\cite{yolo} or SAM~\cite{SAM} can also be used for these tasks. 

A more difficult problem is that of reconstructing \textit{3D information} from sports footage. While some works assume access to multiple camera feeds~\cite{sampleEfficientRL, google} or RGB-D camera feed~\cite{3dBallRecon1}, many do not make this assumption. For instance, ~\cite{sportscameracalib} uses pre-defined court models for camera calibration,~\cite{zhang2023vid2player3d} uses open-source models to reconstruct player joint positions,~\cite{3dBallReconBasketball, 3dBallReconBadmington} reconstruct 3D ball trajectories by using simplistic projectile models, and \cite{Zhang2023Recognizing} reconstructs 3D ball trajectories from low frame-rate footage. Very few works  have combined these methods to provide a simultaneous reconstruction of the player and ball from monocular video. A series of papers in the mid-2000s approached this problem for soccer gameplay~\cite{recon-soccer1, recon-soccer2, recon-soccer3}. To the best of our knowledge, no one has attempted this for table tennis.

\medskip \noindent \textbf{Anticipation in Table Tennis Robotics}\,\,\,\, 
Recent works have designed learning-based systems for various tasks in table tennis robotics \cite{isim2real,google,sampleEfficientRL,Gao2020,Huang2015,Ding2022}, but few have incorporated anticipation into their decision-making algorithm. \cite{janpeters, prob_modeling_of_human_movements} model the unknown human intent using IDDM (Intention-Driven Dynamics Models), a latent variable model, and~\cite{janpeters} uses this information to improve the table tennis playing ability of a simulated robot. \cite{futurepong} trains an LSTM to forecast the landing point of a serve. These works all use relatively small datasets (less than 1000 table tennis exchanges) of gameplay collected in-house. We differ substantially from these works by leveraging a large dataset of expert gameplay.

\medskip \noindent \textbf{Uncertainty Estimation}\,\,\,\, 
Quantifying uncertainty in anticipatory predictions is crucial for deploying a reliable table tennis policy \cite{janpeters}. The primary tools for performing uncertainty estimation with neural networks are ensemble-based methods~\cite{rahaman2021uncertainty, dolezal2022uncertainty, torfah2023learning}, which combine outputs from multiple independently trained models to reduce variance and improve robustness, and conformal prediction~\cite{conformal-pred, angelopoulos2020uncertainty}, which provides calibrated confidence intervals around predictions with user-defined error rates. We combine both methods in this work.

\section{Reconstructing Table Tennis Gameplay}
\label{sec:reconstruction}

In this section, we present our system for reconstructing 3D gameplay from monocular videos of table tennis matches. In particular, we reconstruct the players' 3D SMPL mesh \cite{smpl} and the ball's 3D trajectory, all relative to the world frame (\cref{fig:coordinate_system}). The system comprises three components: video clipping (\cref{subsec:video-clip}), entity tracking (\cref{subsec:entity-track}), and global positioning (\cref{subsec:global-pos}). 



\subsection{Naming Convention}
\label{subsec:naming-convention}



\smallskip \noindent \textbf{Point}\,\,\,\,  A sequence of exchanges. A point starts with a serve and ends when the ball is first out of play.

\smallskip \noindent \textbf{Exchange}\,\,\,\,  Starts and ends with the same player hitting the ball. Exchanges are defined with respect to the player that starts the exchange (we call this player the \textit{ego} of the exchange). One exchange is two segments long. In one exchange, we have a pre-hit segment (time before the opponent hits the ball) and a post-hit segment (time after the opponent hits the ball).

\smallskip \noindent \textbf{Segment}\,\,\,\,  Starts when one player hits the ball and ends when the other player hits the ball. Segments partition the point into disjoint time intervals.

\smallskip \noindent \textbf{Hitting plane}\,\,\,\,  A player's "hitting plane" is the $yz$-plane along the table's edge that is closest to the player.

\subsection{Video Clipping}
\label{subsec:video-clip}

Out of $\sim$800 hours of raw table tennis footage collected from online competition recordings, only $\sim$50 hours of video exhibit actual gameplay. The remaining footage consists of the players walking around, preparing to serve, or replays of prior gameplay. We isolate the gameplay footage with a custom CNN, which is trained to classify frames as occurring \textit{during a point} or \textit{not during a point}. 




\subsection{Entity Tracking}
\label{subsec:entity-track}

The entity tracking component identifies and tracks the key elements in each frame: the table, the rackets, the two players, and the ball. 

\smallskip \noindent \textbf{Table and Rackets}\,\,\,\, 
A YOLOv8 2D image segmentation model~\cite{yolo} segments the table surface, table base, and player rackets, providing masks $M_{\text{table}, t}$, $M_{\text{base}, t}$, and $M^{(i)}_{\text{racket}, t} \text{ for } i \in \{0,1\}$, respectively in the image plane. Note that we do not reconstruct racket poses -- $M^{(i)}_{\text{racket}, t}$ is only used to facilitate the ball reconstruction. 

\smallskip \noindent \textbf{Table Corners}\,\,\,\, High-quality detection of table corners is essential for camera calibration. We first crop the table from each frame using the table mask, $M_{\text{table}, t}$. Next, this is fed into a custom UNet model~\cite{unet} that outputs probable locations of the four table corners and two horizontal table midpoints (the points where the net intersects the top and bottom edges of the table as shown in \cref{fig:ground_projections}). 



\smallskip \noindent \textbf{Players}\,\,\,\, We use HMR 2.0~\cite{hmr2}, a human pose estimator and tracker, to obtain 3D SMPL joint positions for each player relative to the camera coordinate system. We determine the handedness of each player using the location of the racket masks and the pixel coordinates of the players' wrists given by HMR 2.0. 

\smallskip \noindent \textbf{Ball}\,\,\,\, We adopt the TrackNetV3 architecture~\cite{tracknetv3}, fine-tuned on a dataset of 10,000 manually labeled frames, for 2D ball tracking.

\subsection{Global Positioning}
\label{subsec:global-pos}

This stage positions the ball and each player's SMPL mesh in the world frame. 

\begin{figure}[t]
    \centering
    \includegraphics[width=0.7\linewidth]{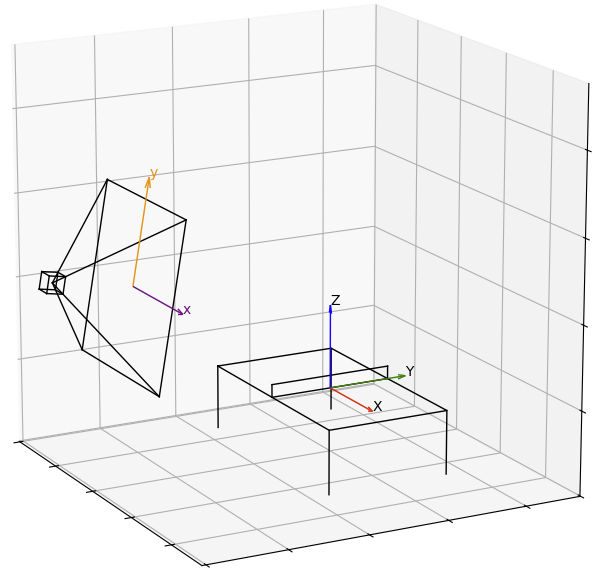}
    \caption{Visualization of the world frame coordinate system (red-green-blue) and image plane coordinate system (purple-orange).}
    \label{fig:coordinate_system}
    \vspace{-4mm}
\end{figure}

\subsubsection{Assumptions}
\label{sec:camera-assumptions}

We assume the following about the camera's pose:

\begin{enumerate}
    \item The camera is translated along the $y$ and $z$ axes only.
    \item The camera's rotation is about the $x$-axis only (see Figure \ref{fig:coordinate_system}) and that the magnitude of this rotation is small enough so that the table's legs are approximately parallel when projected on the image plane. 
\end{enumerate}
We also assume that the table top's width (along the $y$-axis) and the table base's width are the same. These assumptions hold for typical table tennis videos. We filter out videos that do not align with these assumptions during our data collection process. 

\subsubsection{Camera Calibration}

\begin{figure}[t]
    \centering
    \includegraphics[width=0.85\linewidth]{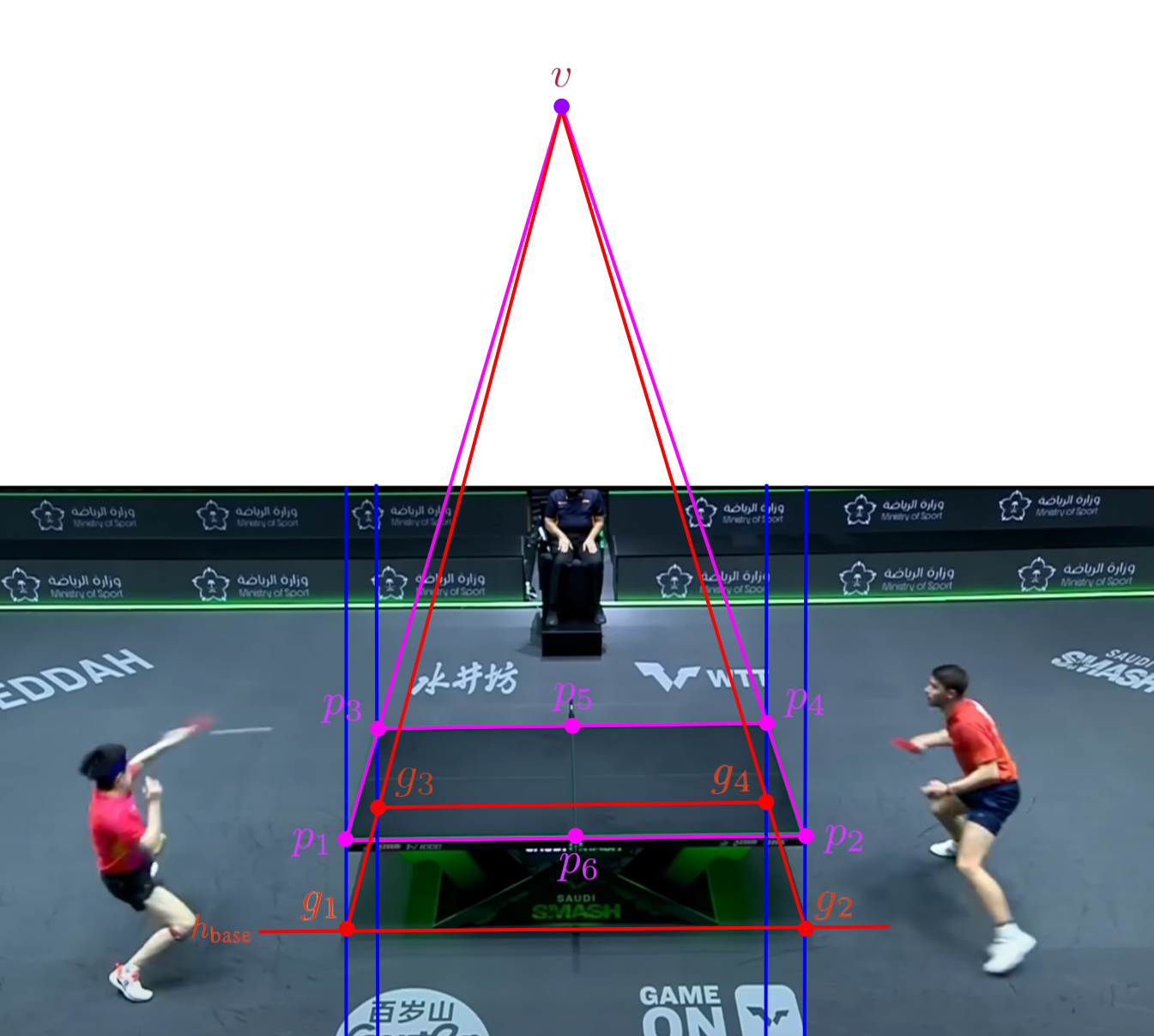}
    \caption{Points used for camera calibration. Ground projection points (in red) are obtained from the table surface points (in pink) by using vanishing points and the table base height.}
    \label{fig:ground_projections}
    \vspace{-2mm}
\end{figure}

We perform camera calibration using the six detected key points on the table surface and the projections of table corners onto the ground plane. Specifically, let $\{ p_{i} \}_{i=1}^6$ denote the six key points returned by the corner detection module at frame $t$ (see \cref{fig:ground_projections}). We define the image height of the table's base as $h_{\text{base}} := \min_{(x,y) \in M_{\text{base}}} \{ y \}$ where $(x,y) \in M_{\text{base}}$ are points in the image plane. The image plane coordinate system we use is visualized in \cref{fig:coordinate_system}.

Our assumptions about the camera's pose and the table base's width allow us to project the two table corners closest to the camera, $p_{1}$ and $p_{2}$, onto the ground plane by drawing vertical lines downward in the image plane from the detected points until they intersect the horizontal line $y = h_{\text{base}}$ (see ~\cref{fig:ground_projections}). Denote these points by $g_{1}$ and $g_{2}$.

To project the two corners furthest from the camera ($p_{3}$ and $p_{4}$) onto the ground plane, we utilize the vanishing point $v$ associated with the lines parallel to the table's depth. Using this, we draw vertical lines downward from $p_{3}$ and $p_{4}$ until they intersect the lines extending from $g_{1}$ and $g_{2}$ to $v$ (see \cref{fig:ground_projections}). 

After this procedure, we obtain a total of 10 image points for calibration. The table dimensions in the competition videos are fixed by ITTF standards \cite{ITTF2024}, so we may use these 10 points to estimate the camera intrinsic matrix $\mathbf{K}$ and extrinsic parameters $(\mathbf{R}, \mathbf{t})$ using standard calibration techniques assuming no camera distortion ~\cite{opencv}.

\subsubsection{Player Positioning}

The 3D SMPL joint positions of the players generated by the entity tracking stage are initially defined in the camera coordinate system. Our positioning procedure draws inspiration from ~\cite{zhang2023vid2player3d}. Instead of using the camera translation $\mathbf{t}$, we compute the 2D position of the player's root projected onto the ground (center of the two ankle keypoints), and then transform this location into world frame coordinates using inverse camera projection. To determine the player's orientation we directly use the calibrated camera rotation matrix $\mathbf{R}$. 

\subsubsection{Ball Trajectory Reconstruction}
\label{sec:ball_recon}

Reconstructing the ball's 3D trajectory from the 2D keypoints involves several steps. This procedure is outlined in \cref{alg:ball_trajectory}.

\begin{algorithm}[t]
\caption{Ball Trajectory Reconstruction}
\label{alg:ball_trajectory}
\begin{algorithmic}[1]
\Require 2D ball positions $\{ b_{\text{2D}, t} \}$, camera intrinsic parameters $\mathbf{K}$ and extrinsic parameters $\mathbf{R}$, $\mathbf{t}$
\State Find hit times $\{ h_i \}_{i=1}^H$
\For{each $i = 1$ to $H-1$}
    \State Find potential bounce times $\{ b_j \}_{j=1}^B \subset [h_i, h_{i+1}]$.
    \For{each $j = 1$ to $B$}
        \State Fit parabolas to $\{ b_{\text{2D}, t} \}_{t=h_i}^{b_j}$ and $\{ b_{\text{2D}, t} \}_{t=b_j}^{h_{i+1}}$
        \State Compute $\text{MSE}_j$ for each fit.
    \EndFor
    \State Select $j^* \in \arg\min_{j \in [B]}\{ \text{MSE}_j \}$ and set $b = b_{j^*}$.
    \State Set $b_{\text{3D}, h_i}$ to player's racket hand at frame $h_i$.
    \State Set $b_{\text{3D}, h_{i+1}}$ to player's racket hand at frame $h_{i+1}$.
    \State Compute $b_{\text{3D}, b}$ via inverse camera projection.
    \State Fit $b_{\text{3D}, t}$ for $t \in [h_i, h_{i+1}]$ via \cref{eq:ball-traj}--(\ref{eq:trajectory_optimization}).
\EndFor
\end{algorithmic}
\end{algorithm}

\smallskip \noindent \textbf{Hit Point Identification}\,\,\,\,
\label{sec:hit-point}
A \textit{hit point} is a timestep $t$ at which either player's racket makes contact with the ball. We identify hit points by detecting local minima in the distance over time between the 2D ball positions $b_{\text{2D}, t}$ and the centroid of the racket masks $M^{(i)}_{\text{racket}, t}$ for each player.

\smallskip \noindent \textbf{Bounce Point Identification}\,\,\,\,
\label{sec:bounce-point}
A \textit{bounce point} is a timestep $t$ at which the ball bounces on the table. Between pairs of hit points, we identify potential bounce points by detecting local minima in the vertical motion of the ball in the image plane. Let $h_1$ and $h_2$ denote a pair of hit points, and let $\{ b_i \}_{i=1}^n$ denote $n$ candidate bounce points between them. For each candidate $b_i$, we fit parabolas to the vertical positions of the ball between frames $h_1$ and $b_i$, and between frames $b_i$ and $h_2$. We compute the mean squared error (MSE) of these fits and select the $b_i$ that minimizes the total MSE:
\begin{equation}
\label{eq:traj-mse}
\text{MSE}(b_i) = \sum_{t = h_1}^{b_i} (y_t - \hat{y}_t^{(1)})^2 + \sum_{t = b_i}^{h_2} (y_t - \hat{y}_t^{(2)})^2,
\end{equation}
where $\hat{y}_t^{(1)}$ and $\hat{y}_t^{(2)}$ are the fitted quadratic functions for the respective intervals. For serves hits, we search over pairs of potential bounce points by fitting 3 parabolic segments instead of 2.

\smallskip \noindent \textbf{Trajectory Fitting}\,\,\,\,
\label{sec:trajectory_fitting}
Since the racket's 3D pose is unknown, at hit points, we approximate the ball's 3D position as the center of the player's racket hand (for which the 3D position is known). At bounce points, we use the fact that the ball's global height is the same as that of the table and use inverse camera projection to determine its 3D position.

Between each pair of hit and bounce points, we model the ball's 3D trajectory with a simplistic projectile model similar to ~\cite{3dBallReconBadmington, 3dBallReconBasketball}. In particular, we model the ball as undergoing projectile motion with Stokes drag \cite{Thornton2007}. Given fixed initial and final positions $b_0 = (x_0, y_0, z_0)$ and $b_T = (x_T, y_T, z_T)$ at times $t = 0$ and $t = T$ (between hit and bounce points), the analytic solution for the ball's trajectory is given by:
\begin{align}
\label{eq:ball-traj}
x_k(t) &= x_0 + (x_T - x_0) \cdot \frac{1 - e^{-k t}}{\,\,1 - e^{-k T}}, \\
y_k(t) &= y_0 + (y_T - y_0) \cdot \frac{1 - e^{-k t}}{\,\,1 - e^{-k T}}, \\
z_k(t) &= z_0 + \left( z_T - z_0 + \frac{g}{k}T \right) \cdot \frac{1 - e^{-k t}}{\,\,1 - e^{-k T}} - \frac{g}{k} t,
\label{eq:ball-traj-end}
\end{align}
where $g$ is the acceleration due to gravity and $k$ is the air resistance coefficient. We optimize $k$ by minimizing the reprojection error:
\begin{equation}
  \min_{k > 0} \sum_{t = 0}^{T} \left\| b_{\text{2D}, t} - \pi(\mathbf{K}, \mathbf{R}, \mathbf{t}, (x_k(t), y_k(t), z_k(t))) \right\|^2.
  \label{eq:trajectory_optimization}
\end{equation}

\subsection{Reconstruction Results}

We applied our reconstruction procedure to $\sim$800 hours of raw table tennis footage. The video clipping stage distills the raw footage down to $\sim$50 hours of gameplay. To ensure robustness of our reconstructions, we discard any points which contain a frame for which the entity tracking stage fails to identify the players, ball, or table. We also discard points for which the MSE in \cref{eq:traj-mse} is above some calibrated threshold. This eliminates almost half of the game-play footage we process. From this, we recovered 73,222 exchanges corresponding to 26.72 hours of pure table tennis gameplay. An example of a reconstructed frame is shown in \cref{fig:reconstruction_example}. 

\subsubsection{Reconstruction Quality}

To assess the accuracy of our reconstruction, we computed the reprojection errors across all frames in our dataset. Note that our videos have $640 \times 360$ pixel resolution. \cref{fig:reproj-errors} plots the reprojection errors for both the recovered ball trajectory and human pose trajectory signals. On average, 1.41 cm around the table's center corresponds to 1 px in the image plane. This implies that our estimate for the ball position and player SMPL joint positions is off on average by $8.9$ cm and $28$ cm, respectively. 


\subsubsection{Dataset Variety and Bias}
The average ball speed in our dataset is $11.25$ m/s, with $80\%$ of speed's falling within the interval $[5.36, 18.27]$ m/s. The average time between hits is $0.56$ seconds.

\smallskip \noindent \textbf{Missing gameplay}\,\,\,\, For each point, our system may not include the last segment if the losing player does not make contact with the ball. This is because the ball reconstruction algorithm requires the ball to be hit by a racket following a bounce point. This segment is arguably the most important portion of a point, as it exhibits the winning move. Failing to reconstruct this critical segment introduces a significant bias in our dataset distribution. 

\begin{figure*}[t]
    \centering
    \begin{minipage}{0.34\textwidth}
        \centering
        \includegraphics[width=\linewidth]{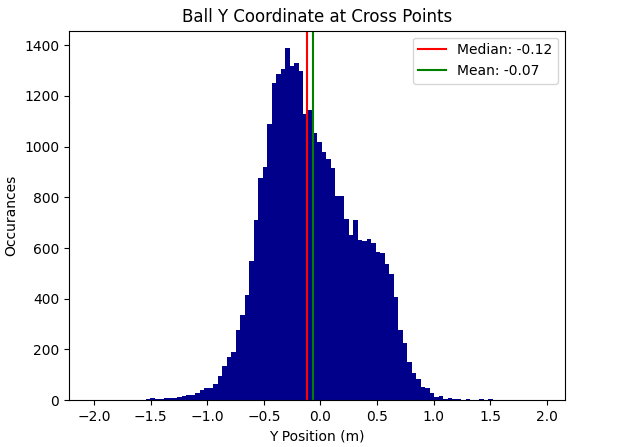}
        \caption{Ball $y$ coordinate at hitting plane}
        \label{fig:ball_y_at_cross_points}
        \vspace{-3mm}
    \end{minipage}%
    \hfill
    \begin{minipage}{0.64\textwidth}
        \centering
        \begin{minipage}{0.49\linewidth}
            \centering
            \includegraphics[width=\linewidth]{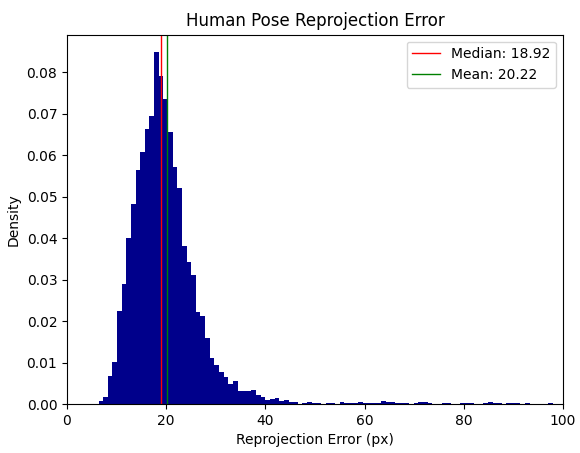}
        \end{minipage}%
        \hfill
        \begin{minipage}{0.49\linewidth}
            \centering
            \includegraphics[width=\linewidth]{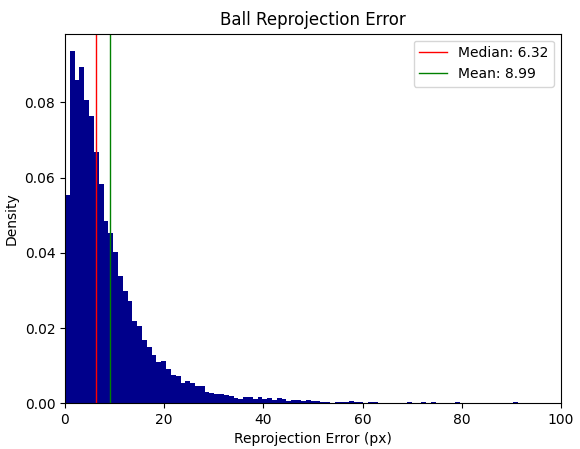}
        \end{minipage}
        \caption{Histograms of ball and player joint position reprojection errors.}
        \label{fig:reproj-errors}
        \vspace{-3mm}
    \end{minipage}
\end{figure*}

\smallskip \noindent \textbf{Population-level Bias}\,\,\,\, We observe what may be a common trend in table tennis gameplay \cite{hitbiases}: players tend to hit the ball towards their right. In particular, the distribution of $y$-positions of the ball trajectory when intersected with the hitting plane appears to be bi-modal (see \cref{fig:ball_y_at_cross_points}). This means that players tend to hit away from the center of the table. This population-level bias is captured in our dataset distribution, and it may influence the performance of our learned models. 


    

\section{Anticipatory Prediction}
\label{sec:anticipatory_prediction}

The ability to anticipate an opponent's intent distinguishes amateur players from professional competitors. This anticipation skill may enable robot systems to play fast-paced, competitive table tennis games, despite fundamental hardware constraints. In this section, we introduce an approach to predict an opponent's intent before they hit the ball. We achieve this by training a generative model—a transformer-based neural network \cite{transformer}—on our extensive dataset of reconstructed table tennis matches. Recent works have shown that transformers can learn complex dynamics and reasoning from large datasets \cite{gpt, hmr2}. We aim for our model to similarly capture the game dynamics and player behaviors in table tennis. By sampling from this model, we generate predictions of ball trajectories based on observed gameplay. We apply conformal prediction to pair these predictions with a measure of uncertainty. 

\subsection{Transformer-based Prediction Model}
\label{subsec:transformer_model}

Let $\mathcal{D}$ denote the dataset of reconstructed points. Each reconstructed point $r = (o_1, o_2, \dots, o_N) \in \mathcal{D}$ consists of a sequence of reconstructed frames $o_i$, which include the opponent's 3D SMPL joint positions, the ego's root position, and the ball's position (all in the world frame).

We tokenize each point into $t = (t_1, t_2, \dots, t_N)$, where each token $t_i$ is formed by concatenating the opponent's SMPL joint positions, the ego's root position only, and the ball's position. Our goal is to model the joint probability density function $p(t)$ auto-regressively:
\begin{equation}
    p(t) = \prod_{i=1}^N p(t_i \mid t_{i-1}, \dots, t_{1}).
\end{equation}
Following the approach in \cite{humanoidasnexttokenpred}, we assume a Gaussian distribution with constant variance and train a transformer to minimize the negative log-likelihood loss $\mathcal{L}$, which takes the form
\begin{equation}
\label{eq:transformer-loss}
    \mathcal{L} = \sum_{t\in \mathcal{D}} \| \hat{t} - t \|^2_2.
\end{equation}
where $\hat{t}$ are the predicted tokens from the transformer.

\subsubsection{Model Architecture}

We model the joint probability density $p(t)$ using a vanilla decoder-only transformer~\cite{transformer} to capture temporal dependencies in the data.  We set the embedding dimension $d = 256$ and use $L = 4$ transformer layers. Each attention layer employs 16 heads, and the MLP has a hidden dimension of 1024. The final model contains approximately 3.2 million learnable parameters, allowing for an average inference time under 10 milliseconds on an Nvidia GeForce RTX 3090 GPU. We train the model for 200 epochs to minimize the loss function in \cref{eq:transformer-loss}.


\subsection{Conformal Prediction}
\label{subsec:conformal_prediction}

We sample from our generative model to obtain predictions of future ball trajectories. To pair these predictions with a measure of uncertainty, we use conformal prediction \cite{conformal-pred}, a statistical framework that allows us to construct prediction intervals with formal guarantees on their coverage probability. 

\subsubsection{Conformal Prediction Theory}

We first provide a brief review of the key result we will use. Suppose we have two datasets, $\mathcal{D}_{\text{train}}$ and $\mathcal{D}_{\text{cal}}$, both consisting of i.i.d. samples $(X_i, Y_i)$ drawn from some unknown distribution on $\mathcal{X} \times \mathbb{R}$. Let $\hat{f}: \mathcal{X} \to \mathbb{R}$ be an estimator for $Y_i$ conditioned on $X_i$ and let $\hat{\sigma}: \mathcal{X} \to \mathbb{R}$ be an estimator of the spread of our predictions. Suppose that both $\hat{f}$ and $\hat{\sigma}$ are independent of the data in $\mathcal{D}_{\text{cal}}$. For each $(X_i, Y_i) \in \mathcal{D}_{\text{cal}}$, we compute the residual 
\vspace{-2mm}
\begin{equation}
    \label{eq:residuals}
    R_i = \frac{|Y_i - \hat{f}(X_i)|}{\hat{\sigma}(X_i)}.
\end{equation}
Letting $n = |\mathcal{D}_{\text{cal}}|$ and given an error tolerance $\alpha \in (0, 1)$, we compute the conformal quantile:
\begin{equation}
    \hat{q}_\alpha = \text{Quantile}\left(\frac{\lceil (n + 1)(1 - \alpha) \rceil}{n}, \frac{1}{n} \sum_{i=1}^{n} \delta_{R_i} \right),
\end{equation}
where $\delta_{R_i}$ is the Dirac measure centered at $R_i$. Then 
setting 
\begin{equation}
\label{eq:conformal-confidence-interval}
\mathcal{C}_\alpha(X) = [\hat{f}(X) - \hat{q}_\alpha\hat{\sigma}(X),\, \hat{f}(X) + \hat{q}_\alpha\hat{\sigma}(X)],
\end{equation} we have the theoretical guarantee $\Pr(Y \in \mathcal{C}_\alpha(X)) \geq 1 - \alpha$.

\subsubsection{Implementing Conformal Prediction}
\label{subsubsec:implementing_conformal_prediction}

To implement conformal prediction, we partition our dataset $\mathcal{D}$ into a training set $\mathcal{D}_{\text{train}}$ and a calibration set $\mathcal{D}_{\text{cal}}$. We further divide $\mathcal{D}_{\text{train}}$ into $k$ disjoint subsets and train an ensemble of $k$ transformers, $T^{(1)}, T^{(2)}, \dots, T^{(k)}$, independently on these subsets. For each $j = 1, \dots, k$ define $\hat{f}^{(j)}(X) := \text{ball's } x\text{-coordinate in } [T^{(j)}(X)]_t$ (\eg the ball's $x$-coordinate in the $t$th token outputted by the transformer). 

For every hit in the $i$th reconstruction $r_i = (o_{i,1}, \dots, o_{i,N}) \in \mathcal{D}_\text{cal}$, we set $X_i = (o_{i,1}, \dots, o_{i,L})$ and $Y_{i} = x_{i,t}$, where $L$ is the timestep some fixed amount before the opponent hits the ball ($L < N$), $t$ is a timestep some fixed amount after the opponent hits the ball, and $x_{i,t}$ is the ball's ground-truth $x$-coordinate in frame $o_{i,t}$. Using our ensemble of transformers, we feed $X_i$ as context and generate $k$ predictions $\hat{f}^{(1)}(X_i), \hat{f}^{(2)}(X_i), \dots, \hat{f}^{(k)}(X_i)$ for the ball's $x$-coordinate at frame number $t$. We then define the aggregated prediction and standard deviation as:
\begin{equation}
    \hat{f}(X) = \frac{1}{k} \sum_{j=1}^{k} \hat{f}^{(j)}(X),
\end{equation}
\begin{equation}
    \hat{\sigma}(X) = \sqrt{ \frac{1}{k} \sum_{j=1}^{k} \left(  \hat{f}^{(j)}(X) - \hat{f}(X) \right)^2 }.
\end{equation}
Using these, we compute the residuals from \cref{eq:residuals} on $\mathcal{D}_\text{cal}$ and construct the conformal confidence intervals $\mathcal{C}_{\alpha,x}(X; t)$ from \cref{eq:conformal-confidence-interval}.

We repeat this process for the ball's $y$-coordinate and $z$-coordinate, constructing two more confidence intervals $\mathcal{C}_{\alpha,y}(X; t)$ and $\mathcal{C}_{\alpha,z}(X; t)$. Finally, we form the confidence region $\mathcal{C}_{\alpha}(X; t) = \mathcal{C}_{\alpha,x}(X; t) \times \mathcal{C}_{\alpha,y}(X; t) \times \mathcal{C}_{\alpha,z}(X; t)$. Assuming samples are drawn i.i.d, an application of the union bound shows that the ball's true position $b_t = (x_t, y_t, z_t)$ at frame number $t$ satisfies:
\begin{equation}
    \Pr(b_t \in \mathcal{C}_\alpha(X; t)) \geq 1 - 3\alpha.
\end{equation}

\section{Experiments and Results}
\label{sec:exp-res}

In this section, we evaluate the effectiveness of our model for making anticipatory predictions about an opponent's actions in table tennis. We conduct a series of experiments to demonstrate the utility of our dataset in improving robot performance in fast-paced gameplay.

\subsection{Experimental Setup}

We partitioned our dataset $\mathcal{D}$ into five disjoint training subsets $\mathcal{D}_1, \dots, \mathcal{D}_5$, a calibration set $\mathcal{D}_{\text{cal}}$ of 2,500 exchanges, and a test set $\mathcal{D}_{\text{test}}$ of 1,000 exchanges. We trained an ensemble of five transformers on the training subsets. The calibration set was used to compute the conformal quantile $\hat{q}_\alpha$ for various significance levels $\alpha$. We then ran the following experiments on the held-out test set $\mathcal{D}_{\text{test}}$.

\subsection{Evaluating the Confidence Regions}
\label{subsec:evaluation}

We first verify that the conformal prediction intervals satisfy the theoretical coverage guarantees. For each exchange in the test set, we compute the conformal regions $\mathcal{C}_{\alpha}(X; t)$ for $\alpha = 0.10$, $0.15$, and $0.20$. To evaluate the coverage, we measured the proportion of times the true ball position fell within the predicted confidence intervals along each axis and within the entire confidence region $\mathcal{C}_{\alpha}(X; t)$. \Cref{tab:coverage_results} summarizes these results. We see that the empirical coverage closely matches the theoretical guarantees along each axis. We also observe that the total coverage exceeds the theoretical lower bound, implying that the union bound is loose in this setting.

\vspace{-2mm}

\begin{table}[ht]
\centering
\begin{tabular}{cccccc}
\toprule
$1 - \alpha$ & $\mathcal{C}_{\alpha, x}$ & $\mathcal{C}_{\alpha, y}$ & $\mathcal{C}_{\alpha, z}$ & $1 - 3\alpha$ & $\mathcal{C}_{\alpha}$ \\
\midrule
0.90 & 0.885 & 0.906 & 0.905 & 0.70 & 0.763 \\
0.85 & 0.830 & 0.858 & 0.862 & 0.55 & 0.652 \\
0.80 & 0.782 & 0.796 & 0.822 & 0.40 & 0.532 \\
\bottomrule
\end{tabular}
\caption{Empirical coverage probabilities for different significance levels $\alpha$ on the test set. The theoretical coverage probabilities are given by $1 - \alpha$ and $1 - 3\alpha$.}
\label{tab:coverage_results}
\vspace{-3mm}
\end{table}

\begin{figure}[h]
\centering
\includegraphics[width=0.75\linewidth]{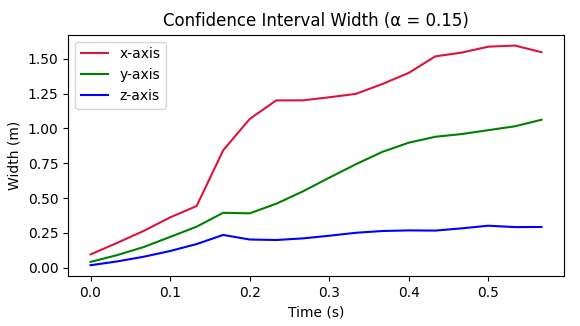}
\caption{Average confidence interval width along each axis as a function of the prediction time horizon.}
\label{fig:ci_width_vs_time}
\vspace{-4mm}
\end{figure}

We also analyzed the average width of the confidence intervals along each axis at different prediction timesteps $t$ (see Figure~\ref{fig:ci_width_vs_time}). We observed that as the prediction horizon increases, the confidence intervals become wider, reflecting increased uncertainty in the predictions. We also see that the confidence intervals differ in size along each axis. For instance, the confidence intervals are much larger along the x-axis than along the z-axis. This is explained by the fact that the ball's $x$-axis speed is usually larger than its $y$-axis and $z$-axis speeds.  
\vspace{-4mm}

\subsubsection{Analysis of Extreme Hits}

We define \textit{extreme hits} as opponent hits which cross the ego player's hitting plane with an absolute $y$-value greater than $0.75$ meters (i.e. hits to the far left or far right side of the table). For each extreme hit in the test set, we examined whether the confidence intervals were biased toward the correct direction at the moment the ball crosses the robot's hitting plane. In particular, we say that a confidence region is \textit{biased left} if it excludes at least $1/3$ of the right side of the table. We define \textit{biased right} analogously. The analysis shows that the confidence regions usually over-cover the table on extreme hits. However, of the subset of confidence regions that are biased, they are biased towards the correct side more than $85\%$ of the time (see Table~\ref{tab:extreme_hits_bias}).

\begin{table}[ht]
\centering
\begin{tabular}{lcc}
\toprule
           & True Right & True Left \\
\midrule
Over Covered  & 0.814 & 0.583 \\
Biased Right  & \textbf{0.163} & 0.039 \\
Biased Left   & 0.023 & \textbf{0.388} \\
\bottomrule
\end{tabular}
\caption{Proportion of extreme hits where the confidence intervals were biased in the correct direction at the hitting plane ($\alpha = 0.15$).}
\label{tab:extreme_hits_bias}
\vspace{-3mm}
\end{table}

\subsection{Anticipatory Control Algorithm}
\label{subsec:control_algorithm}

To test the utility of the conformal confidence regions, we designed a simple control algorithm that enables the robot to preposition itself before an opponent hits the ball.

\subsubsection{Robot Reachable Set Modeling}

We model the robot arm's reachable set $\mathcal{R}(t)$ as growing linearly over time, starting from the current end-effector position $p_0 \in \mathbb{R}^3$. Specifically, we use the model:
\begin{equation}
    \mathcal{R}(t) = \left\{ p \in \mathbb{R}^3 \, \big| \, \left\| p - p_0 \right\|_2 \leq v_{\text{max}} t \right\} \cap \mathcal{B},
    \label{eq:reachable_set}
\end{equation}
where $v_{\text{max}}$ is a fixed maximum velocity, and 
$\mathcal{B}$ is the precomputed safe, reachable set determined by the environment and the robot's kinematics. 

\subsubsection{Target Confidence Region Selection}

Let $X = (o_1, \dots, o_L)$ be a sequence of reconstructed frames of a point, ending just before the opponent hits the ball. As described in \cref{subsubsec:implementing_conformal_prediction} we generate the conformal confidence region $ \mathcal{C}_{\alpha}(X; t)$ for each $t > L$.  We then set, 
\begin{equation}
    t^\star = \min\{t : \mathcal{C}_{\alpha}(X; t) \subseteq \mathcal{R}(t - L) \}.
\end{equation}

\subsubsection{Target Pre-Positioning Selection}

We would like to select a point $p^\star \in \mathcal{C}_{\alpha}(X; t^\star)$ at which the agent pre-positions its end-effector. A straightforward choice would be the centroid $c_\alpha$ of $\mathcal{C}_{\alpha}(X; t^\star)$. In practice, we find that applying some shrinkage towards a fixed central position leads to better performance. Let $c$ be such a fixed position. We set 
\begin{equation}
    p^\star = \Pi_{\mathcal{C}_{\alpha}} \left( \lambda c + (1 - \lambda) c_\alpha \right)
\end{equation}
where $\Pi_{\mathcal{C}_{\alpha}}$ is the projection mapping onto $\mathcal{C}_{\alpha}(X; t^\star)$ and $\lambda \in [0, 1]$ is a hyper-parameter.

\subsection{Anticipatory Controller Performance}
\label{subsec:control_performance}

We evaluated the effectiveness of our anticipatory control algorithm on a robotic platform in simulation and measured its impact on the ability to return shots. 

\subsubsection{Simulation Setup}

\begin{figure}
    \centering
    \includegraphics[width=0.4\textwidth]{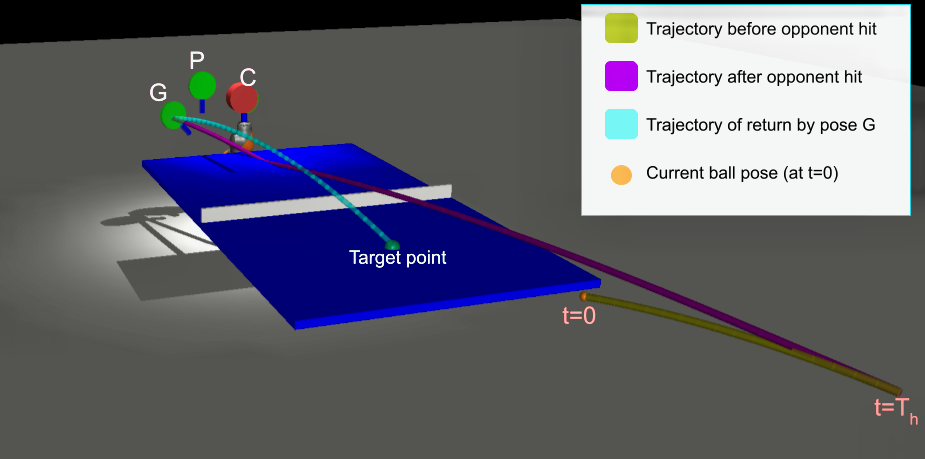}
    \caption{Simulated target poses and ball trajectory.}
    \label{fig:sim_comp}
    \vspace{-4mm}
\end{figure}

\begin{figure}[t]
    \centering
    \includegraphics[width=0.3\textwidth]{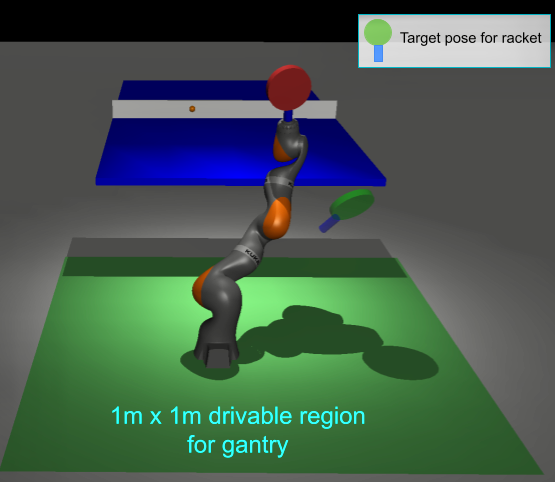} 
    \caption{Simulator snapshot.}
    \label{fig:sim_snap}
    \vspace{-5mm}
\end{figure}

We used a KUKA robotic arm within the MuJoCo physics simulation environment~\cite{mujoco}. The arm is mounted on a gantry with a top speed of $2$ m/s, similar to the setup in~\cite{google}. The design parameters, such as arm size and the gantry's traversable space, also mirror those in~\cite{google}. A snapshot of the environment is shown in ~\cref{fig:sim_snap}. Our control objective is to maximize the \textit{ball return rate}, defined as the proportion of incoming shots the robot successfully returns. To achieve this, we designed a blocking controller where the robot aims to block incoming shots without performing a swing. The target pose for the racket, achieved by the end-effector, is set to return the ball to the center of the opponent's side of the table, as illustrated in \cref{fig:sim_comp}. We note that the controller design is independent of the anticipation algorithm described in \cref{sec:anticipatory_prediction}. We use this setup solely to measure the improvement in performance that anticipation brings in terms of the ball return rate. We do not claim that the controller is optimal; therefore, we also report the return rate when provided with the ground truth trajectory of the ball, treating it as an oracle. This represents the best performance our baseline controller can achieve given perfect information. Improving the controller's performance is a separate research direction explored in~\cite{Gao2020, buchler2022learning} which we do not focus on here. Our aim is to demonstrate that anticipation can enhance the performance of this baseline controller, and a more advanced controller might further improve results with the same trend.

The simulation environment models the dynamics of the robot and the ball, incorporating realistic physical properties. More details on the exact algorithm used in designing the controller are presented in the appendix. We evaluated the robot's performance under three different scenarios:

\begin{enumerate}
    \item \textbf{Baseline (no pre-positioning)}: The robot remains stationary at it's central position (pose C in \cref{fig:sim_comp}) until the opponent strikes the ball, at which point it moves to the position where the ground truth ball trajectory intersects its hitting plane (pose G in \cref{fig:sim_comp}).
    \item \textbf{Anticipatory pre-positioning}: The robot uses our anticipatory control algorithm to pre-position its end-effector at the predicted target point $p^\star$ (pose P in \cref{fig:sim_comp}) before the opponent hits the ball. We fix some $T_h > 0$ and generate $p^\star$ exactly $T_h$ seconds before the opponent hits the ball in each exchange. 
    The target pose is set to G after the opponent hits the ball.
    \item \textbf{Oracle pre-positioning}: The robot is provided with the ground truth future ball trajectory and pre-positions accordingly (pose G in \cref{fig:sim_comp}) exactly $T_h$ seconds before the opponent hits the ball.
\end{enumerate}

\subsubsection{Results}

Table~\ref{tab:return_rates} presents the ball return rates for each scenario over a held out test set of exchanges. These results demonstrate the utility of an anticipatory control algorithm in improving the ball return rate for a robot table tennis player. While the oracle pre-positioning achieves the highest return rate, our predictive approach closes a substantial portion of the gap without requiring perfect foresight. We set $T_h = 0.2s$ and $\alpha=0.15$ for this experiment. Some more additional results and ablation studies are in Appendix D.

\begin{table}[ht]
\centering
\begin{tabular}{lccc}
\toprule
\shortstack{Pre-Pos. \\ Strategy} & \shortstack{Return \\ Rate} & \shortstack{Return \\ Accuracy} & \shortstack{Pose \\ Accuracy} \\
\midrule
Baseline     & 49.9\%         & 0.497 m                     & 0.25 m / 13.3$^{\circ}$  \\
Anticipatory & \textbf{59.0\%} & \textbf{0.463 m}            & \textbf{0.19 m / 9.86$^{\circ}$} \\
Oracle       & \textit{64.5\%} & \textit{0.453 m}            & \textit{0.15 m / 6.26$^{\circ}$} \\
\bottomrule
\end{tabular}
\caption{Return rates for the robot under different pre-positioning strategies. Return accuracy is the mean deviation of the return bounce point from the target point on the opponent's side of the table. Pose accuracy is the mean difference in pose (position error/orientation error) of the racket achieved vs target pose at the time when the ball hits or passes the racket.}
\label{tab:return_rates}
\vspace{-5mm}
\end{table}

\section{Conclusion}
\vspace{-1mm}

We present a scalable method for reconstructing 3D table tennis gameplay from monocular videos, and we leverage publicly available footage to create a large-scale dataset comprising over 73,000 exchanges. This dataset enabled us to train a transformer-based model to anticipate opponent actions, predicting future ball trajectories \textit{before} opponents make contact with the ball. We integrate these predictions into a controller for a KUKA robot arm and demonstrate improved human-robot gameplay in simulation, increasing the ball return rate from from 49.9\% to 59.0\% as compared to a baseline policy. Our work underscores the importance of behavior-aware algorithms in designing human-robot control algorithms. Future work includes reducing reconstruction error, extending the reconstruction algorithm to operate on a larger variety of camera views, addressing dataset biases, and validating the approach on physical hardware. 
\section{Acknowledgements}
\vspace{-1mm}

This research was supported in part by  DOD Advanced Research Projects Agency project Design of Robustly Implementable Autonomous and Intelligent Machines under Award HR00112490425. This research was also supported in part by DAF Air Force Research Laboratory project Provably Correct Design of Adaptive Hybrid Neuro-Symbolic Cyber Physical Systems under Award FA8750-23-C-0080. 

{
    \small
    \bibliographystyle{ieeenat_fullname}
    \bibliography{main}
}

\clearpage
\setcounter{page}{1}
\maketitlesupplementary


\appendix

\section{Design of the controller to track any target pose of the end-effector}

We use reinforcement learning to train a policy that can achieve a target pose within a given distribution in minimum time. The robot consists of manipulator with 7 joints and movement of the base in XY plane resulting into a 9DoF systems as it has 9 control variables to move the system. The racket is attached at the end-effector as shown in Figure \ref{fig:sim_snap}. Reason for choice of a distribution for the target during training is to encourage taking solutions of joint angles and gantry position such that it can easily move to any pose within the distribution and also change within the same distribution very quickly. The exact joint angle range of the Kuka iiwa R820 are given in Table \ref{tab:ang_range}. We now describe the RL algorithm in detail 

\begin{table}[]
\centering
\begin{tabular}{lcc}
\hline
Joint & Min angle (in $^{\circ}$) & Max angle (in $^{\circ}$) \\ \hline
A1    & -170            & 170             \\ 
A2    & -120            & 120             \\
A3    & -170            & 170             \\ 
A4    & -120            & 120             \\ 
A5    & -170            & 170             \\ 
A6    & -120            & 120             \\ 
A7    & -175            & 175             \\ \hline
\end{tabular}
\caption{Kuka IIWA R820 joint angle ranges}
\label{tab:ang_range}
\end{table}

\subsection{Reward design}

We reward the RL agent for moving towards the target pose. Given a target goal $T$. The distance cost from the goal, $C$ is defined as follows:-

\begin{equation}
    C = (ee_\text{position} - T_\text{position})^2 + w_{\text{orientation}} \cosh{((ee_\text{quat} T_\text{quat}^{-1}).w)}
\end{equation}

where $w_{\text{orientation}}$ is the weight factor for the orientation error in radians and is a hyperparameter set to $0.4$ for the experiments in this paper, $T_\text{position}$ is the goal position and $T_\text{quat}$ is the goal orientation in quaternion, $ee_\text{position}$ and $ee_\text{quat}$ are the racket position and quaternion that is attached at the end-effector. The reward, $R_t$ at each time step is given as follows:-

\begin{equation}
\begin{split}
    &R_t = -(C_t - C_{t-1}) + w_\text{ac} |a_t|^2 \text{ for } t>0 \\
    &R_0 = |a_0|^2
\end{split}
\end{equation}

where $a_t$ is the action command at time step $t$ defined as the change in joint angle and gantry position wrt the current configuration. At every time step $t$, the target pose $T_t$ is changed if the robot is able to match the pose of the racket on it's end-effector with some tolerance $\text{tol}_t$. The tolerance $\text{tol}_t = \text{tol}_\text{final} + (\text{tol}_\text{init}-\text{tol}_\text{final}) \exp^{-t/t_\text{half}}$ is changed in a curriculum from $\text{tol}_\text{init}$ to $\text{tol}_\text{final}$. Mathematically, change in target pose $T_t$ is defined as follows:-

\begin{equation}
\begin{split}
&T_t = 
\begin{cases} 
    T_{t-1} & \text{if } C_t \leq \text{tol}_t, \\ 
    T_\text{new} \sim T_\text{dis} & \text{otherwise}
\end{cases} \text{ for } t > 0 \\
&T_0 \sim T_\text{dis}
\end{split}
\end{equation}

\subsection{Training target pose distribution}

\begin{figure}
    \centering
    \includegraphics[width=0.9\linewidth]{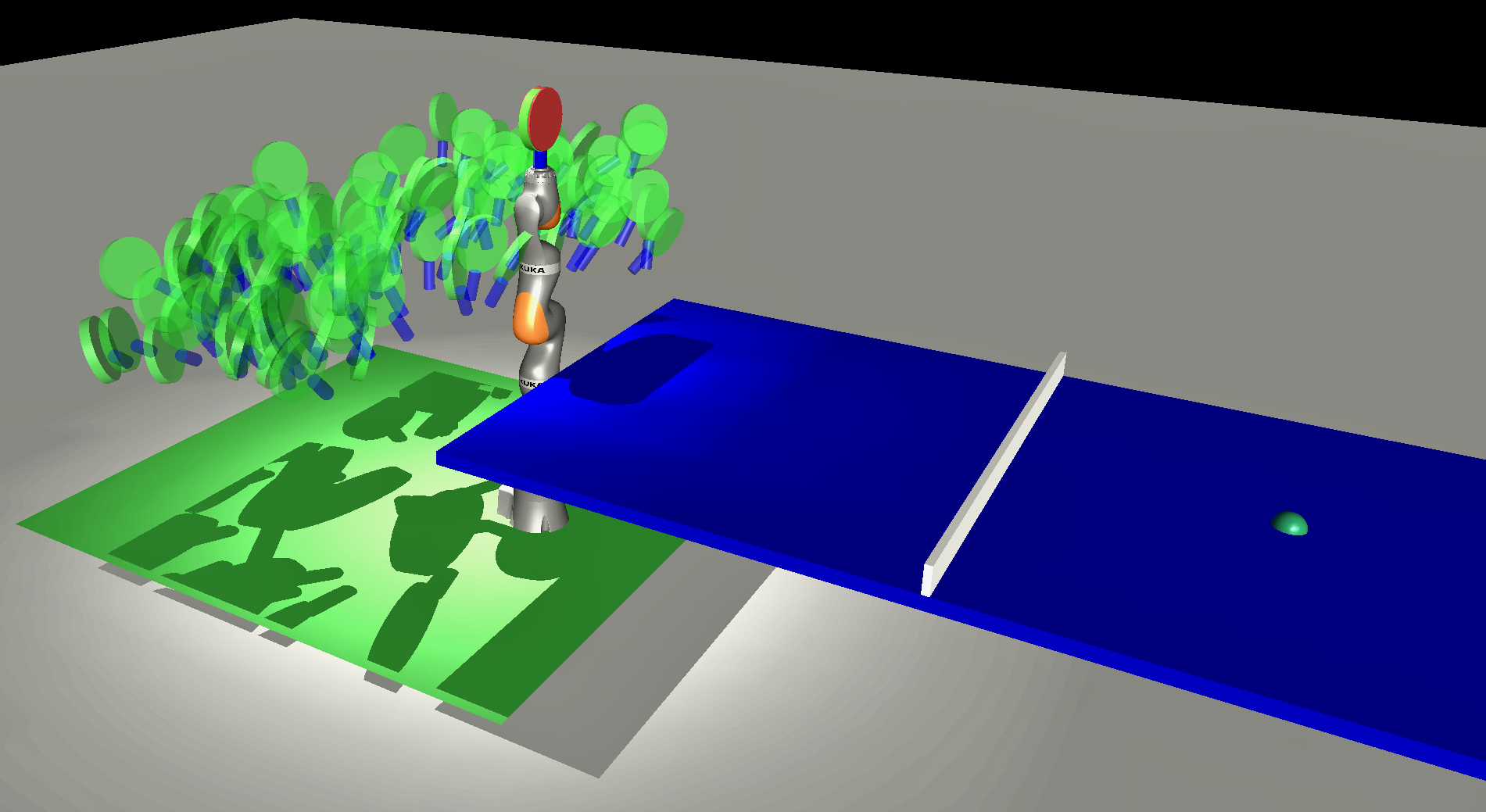}
    \caption{100 poses sampled from $T_\text{dis}$}
    \label{fig:train_dis}
\end{figure}

The target pose distribution $T_\text{dis}$ is defined as follows:-

\begin{equation}
\begin{split}
    T_\text{dis} = &\{ \text{pos} \sim \mathcal{U}(\text{pos}_\text{min},\text{pos}_\text{max}) \\
    &\text{euler} = \text{euler}_\text{facing} + e \sim \mathcal{U}(-\text{euler}_\text{range},\text{euler}_\text{range})\}
\end{split}
\end{equation}

where $\text{euler}_\text{facing}$ is the orientation of racket face that extends to the target point in Figure \ref{fig:sim_comp} with handle facing in -Z direction; $\text{pos}_\text{min},\text{pos}_\text{max},\text{euler}_\text{range}$ are chosen accordingly. This enables learning to achieve poses only within the distribution of where players usually hold their racket to return the ball facing the table. $100$ randomly chosen poses in this distribution are given in Figure \ref{fig:train_dis}

\subsection{Policy training}

\begin{figure}[ht]
    \centering
    \begin{subfigure}{0.23\textwidth}
        \centering
        \includegraphics[width=\textwidth]{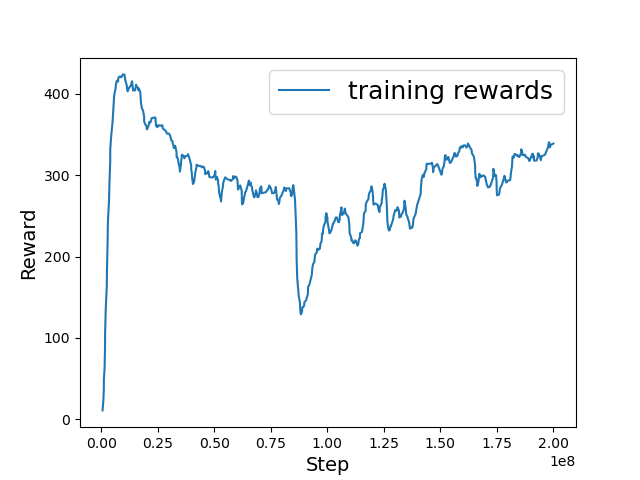}  
        \caption{Train}
        \label{fig:train}
    \end{subfigure}
    \hfill
    \begin{subfigure}{0.23\textwidth}
        \centering
        \includegraphics[width=\textwidth]{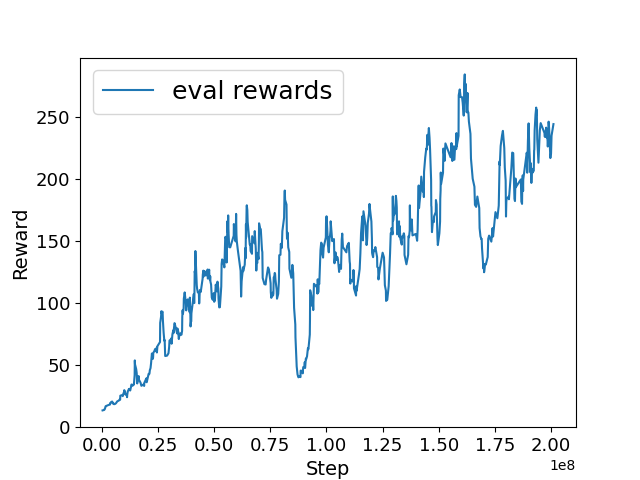}  
        \caption{Eval}
        \label{fig:eval}
    \end{subfigure}
    \caption{Rewards}
    \label{fig:rewards}
\end{figure}

We use Proximal Policy Optimization (PPO) RL algorithm to learn the optimal policy to maximize the reward in the environment described above. The episode sixe is chosen to be $1000$ time steps with a time step $dt=0.01s$. The training curve for rewards and the evaluation performance is given in Figure \ref{fig:rewards}. For evaluation performance, we fix $\text{tol} = \text{tol}_\text{final}$ and choose the policy with best performance within $200M$ environment steps

\section{Contact modeling}

It is important to specify how the ball will bounce when collided with the table and the racket so as to recreate the ball exchanges collected from videos as described in Section \ref{sec:reconstruction}

\subsection{Racket to ball contact} \label{app:racket_to_ball}

For the racket to ball contact we assume lossless bounce i.e. given the racket normal $n_\text{racket}$ and the ball velocity before collision $v_\text{before}$. The velocity after collision, $v_\text{after} = v_\text{before} + 2 (v_\text{before} . n_\text{racket}) n_\text{racket}$. This conserves the speed ($|v_\text{before}| = |v_\text{after}|$) and as the racket is assumed to be static, it does not impart any extra speed to the ball. 

\subsection{Ball to table contact}

For the ball to table contact, we aim to exactly recreate the trajectory parabola obtained in Section \ref{sec:reconstruction}. For this, we fit the ball velocities for each segment divided by when ball contacts the table or the opponent's racket. Whenever ball makes a collision with the table or the opponent's racket, the new velocity is set to the values fit for the new segment as fit in Section \ref{sec:reconstruction}. This very closely enables  recreating the trajectories in Section \ref{sec:reconstruction} 

\section{Setting target pose G} 

The target pose G is calculated as the pose to reflect the ball based on Section \ref{app:racket_to_ball} such that it hits the table as close as possible to the target point $b_\text{target}$. Mathematically, given the table height $z_\text{table}$, velocity of the ball at the hitting point $v_\text{before}$, the hitting point on the racket $b_\text{hit}$, and the function $f_\text{norm}$ that extracts the racket hitting plane normal from pose, this can be formulated as:-

\begin{equation}
\begin{split}
    G = \underset{G}{arg\,min} &|b_\text{target} - p|^2 \\
    \text{where } &p_x = b_{\text{hit},x} + v_{\text{reflected},x} t_c \\
    &p_y = b_{\text{hit},y} + v_{\text{reflected},y} t_c \\
    &v_\text{reflected} = v_\text{before} + (v_\text{before} . f_\text{norm}(G)) f_\text{norm}(G)\\
    &t_c = \frac{-\sqrt{v_{\text{reflected},z}^2 - 2 g (b_{\text{hit},z}-z_\text{table})} - v_{\text{before},z}}{g}
\end{split}
\end{equation}

where $g=-9.81 m/s^2$ is the acceleration due to gravity constant

\section{Additional results} \label{sec:add_results}

\subsection{Varying central pose $C$}

The results presented in Table \ref{tab:return_rates} are with the central position, $C$ at the center. However, we also set pose $C$ not to be in the center of the table but biased towards one side. We set $C$ as the mean of all the positions in the training dataset from where the player hits the ball. This is to show that the transformer does not just learn the mean of all returns made by the opponent but rather a correlation with the current pose, history of poses of the opponent. The updated results are given in Table \ref{tab:return_rates_C}. The return rates improve with the updated starting pose $C$ for all the 3 cases, however the trend remains the same, i.e. the return rates with using the ground truth is the highest, followed by using our prediction for pre-positioning and the return rate if not pre-positioning is the lowest. This shows that the trained transformer indeed learns a correlation of the human poses with the anticipated return trajectory of the ball.  

\begin{table}[ht]
\centering
\begin{tabular}{lccc}
\toprule
\shortstack{Pre-Pos. \\ Strategy} & \shortstack{Return \\ Rate} & \shortstack{Return \\ Accuracy} & \shortstack{Pose \\ Accuracy} \\
\midrule
Baseline     & 52.9\%         & 0.506 m                     & 0.23 m / 12.68$^{\circ}$  \\
Anticipatory & \textbf{62.5\%} & \textbf{0.469 m}            & \textbf{0.17 m / 9.43$^{\circ}$} \\
Oracle       & \textit{66.2\%} & \textit{0.489 m}            & \textit{0.14 m / 5.93$^{\circ}$} \\
\bottomrule
\end{tabular}
\caption{Return rates for the robot under different pre-positioning strategies with $C$ as the mean position of all the hit points in the training dataset and orientation facing the table. Return accuracy is the mean deviation of the return bounce point from the target point on the opponent's side of the table. Pose accuracy is the mean difference in pose (position error/orientation error) of the racket achieved vs target pose at the time when the ball hits or passes the racket.}
\label{tab:return_rates_C}
\end{table}

\subsection{Varying $\lambda$}

Next, we study the effect on results by changing the hyperparameter $\lambda$ which dictates how much to trust the prediction made by the anticipatory algorithm. $\lambda=0$ means trusting the prediction completely, while $\lambda=1$ means the anticipatory algorithm is completely trusted in setting the target pose. Table \ref{tab:return_rates_lambda} shows results with different values of $\lambda$. As can be seen, we get the best return rate at $\lambda=0.1$  

\begin{table}[ht]
\centering
\begin{tabular}{lccc}
\toprule
\shortstack{Pre-Pos. \\ Strategy} & \shortstack{$\lambda=0$} & \shortstack{$\lambda=0.1$} & \shortstack{$\lambda=0.5$} \\
\midrule
Baseline     & 49.9\%         & 49.9\%                     & 49.9\%   \\
Anticipatory & 55.6\%  & \textbf{59.0 \%}            & 54.4\%  \\
Oracle       & \textit{64.5\%} & \textit{64.5\%}            & \textit{64.5\%} \\
\bottomrule
\end{tabular}
\caption{Return rates for the robot under different values of $\lambda$ and using our anticipatory pre-positioning strategy}
\label{tab:return_rates_lambda}
\end{table}

\subsection{Varying $T_h$}

Next, we study the effect on results by changing the hyperparameter $T_h$, the length of history that dictates the time to anticipate before the opponent hits the ball. There is a trade-off between accuracy and the available time for the anticipatory algorithm to pre-position. With larger $T_d$, the anticipatory algorithm will have more response time, but the anticipation will be more inaccurate, as it will be hard to tell how the opponent will hit more ahead of time. Vice-versa for shorter $T_h$. Hence, the value of $T_h$ is chosen accordingly. Table \ref{tab:return_rates_th} shows results with different values of $T_h$. As can be seen, we get the best return rate at $T_h=0.2s$.

\begin{table}[ht]
\centering
\begin{tabular}{lccc}
\toprule
\shortstack{Pre-Pos. \\ Strategy} & \shortstack{$T_h=0.1s$} & \shortstack{$T_h=0.2s$} & \shortstack{$T_h=0.4s$} \\
\midrule
Baseline     & 49.9\%         & 49.9\%                     & 49.9\%   \\
Anticipatory & 57.3\%  & \textbf{59.0 \%}            & 58.4\%  \\
Oracle       & \textit{62.1\%} & \textit{64.5\%}            & \textit{65.2\%} \\
\bottomrule
\end{tabular}
\caption{Return rates for the robot under different values of $\lambda$ and using our anticipatory pre-positioning strategy}
\label{tab:return_rates_th}
\end{table}

\end{document}